\pdfoutput=1

\documentclass[11pt]{article}

\usepackage[]{EMNLP2022}

\usepackage{times}
\usepackage{latexsym}
\usepackage{enumitem}

\usepackage[T1]{fontenc}

\usepackage[utf8]{inputenc}

\usepackage{graphicx}
\usepackage[skip=4pt]{caption}
\usepackage{caption}
\usepackage{subcaption}
\usepackage{xcolor}
\usepackage{multirow}
\definecolor{Mycolor1}{HTML}{004488}
\definecolor{Mycolor2}{HTML}{DDAA33}
\definecolor{Mycolor3}{HTML}{BB5566}
\usepackage{changes}

\usepackage{microtype}

\usepackage{inconsolata}

\definechangesauthor[color=red]{OA}
\definechangesauthor[color=blue]{EC}
\newcommand{\squeezeup}{\vspace{-2.5mm}}

\newcommand\T{\rule{0pt}{2.6ex}}       
\newcommand\B{\rule[-1.2ex]{0pt}{0pt}} 

\title{Cognitive Simplification Operations Improve Text Simplification}

\author{Eytan Chamovitz \\
   Department of Computer Science \\
   Tel Aviv University \\
   \texttt{eytanc@gmail.com} \\
\And
  Omri Abend \\
   Department of Computer Science \\
   Hebrew University of Jerusalem \\
   \texttt{omri.abend@mail.huji.ac.il} \\
  }

\begin{document}
\maketitle
\begin{abstract}
Text Simplification (TS) is the task of converting a text into a form that is easier to read while maintaining the meaning of the original text. A sub-task of TS is Cognitive Simplification (CS), converting text to a form that is readily understood by people with cognitive disabilities without rendering it childish or simplistic. This sub-task has yet to be explored with neural methods in NLP, and resources for it are scarcely available. In this paper, we present a method for incorporating knowledge from the cognitive accessibility domain into a TS model, by introducing an inductive bias regarding what simplification operations to use. We show that by adding this inductive bias to a TS-trained model, it is able to adapt better to CS without ever seeing CS data, and outperform a baseline model on a traditional TS benchmark. In addition, we provide a novel test dataset for CS, and analyze the differences between CS corpora and existing TS corpora, in terms of how simplification operations are applied. 
\end{abstract}

\section{Introduction}\label{sec:intro}

\emph{Text Simplification} (TS) is the task of converting text into a form that is easier to understand by modifying its syntax and/or the words used in it, while maintaining the original text's meaning \citep{Alva-Manchego2020}. 

TS is a very diverse task that can include simplifications aimed at different target audiences. TS is often operationalized in NLP using a number of particular corpora to train and evaluate neural models (see \S\ref{sec:rw}), whose target audiences are mostly second language learners, primary school students or adults with learning disabilities. For brevity, this paper will  refer by TS to this concrete formulation of the simplification task, rather than the abstract, general notion of the task of simplifying text.

\emph{Cognitive Simplification} (CS) is the task of converting text to a form that is clear, simple, and readily understood by people with cognitive disabilities \cite{Yalon-Chamovitz2009A,Yalon-Chamovitz2016,Yalon-chamovitz2016B}.\footnote{People with developmental disabilities, head trauma patients, people with dementia or Alzheimer's Disease, etc. Not including people with learning disabilities such as dyslexia.} 
The procedure includes structural and lexical modifications that reduce the text's complexity, while preserving \emph{as much} of the meaning and information content as possible, and without rendering it childish or simplistic. See \autoref{fig:approach}.

\begin{figure*}[htbp]
    \centering
    \small
     \renewcommand{\arraystretch}{1.5}
    \begin{tabular}{p{0.2\linewidth} p{0.75\linewidth}}
        \textbf{Original Source}: & \textcolor{Mycolor1}{\textbf{Now, normally}} during Disability Pride Month, we're \underline{\textcolor{Mycolor2}{\textbf{showcasing our disability pride}}} through \textit{various} \underline{parades and events \textcolor{Mycolor3}{\textbf{throughout the country}}}. \\
        \textbf{Original Target}: & \textcolor{Mycolor1}{\textbf{Most years}}, during Disability Pride Month we have \underline{parades and events \textcolor{Mycolor3}{\textbf{all over the}}} \underline{\textcolor{Mycolor3}{\textbf{United States}} to \textcolor{Mycolor2}{\textbf{show how proud we are}}}. \vspace{4pt}\\
        \textbf{Operations}: &  <REPHRASE> <DEL> <REORDER> \\
        \hline
        \textbf{Modified Source T5}: &  <mask\_1> Now, normally during Disability Pride Month, we're showcasing our disability pride through various parades and events throughout the country. \\
        \textbf{Modified Target T5}: &  <mask\_1> <REPHRASE> <DEL> <REORDER> <mask\_2> Most years, during Disability Pride Month we have parades and events all over the United States to show how proud we are. \\
        \hline
        \textbf{Modified Source BART}: &  <mask> Now, normally during Disability Pride Month, we're showcasing our disability pride through various parades and events throughout the country. \\
        \textbf{Modified Target BART}: &  <REPHRASE> <DEL> <REORDER> Most years, during Disability Pride Month we have parades and events all over the United States to show how proud we are. \\
    \end{tabular}
    \caption{\small Illustration of our approach on an example sentence from the CS dataset FestAbility Transcripts. The modified sources and targets for each model architecture include special operation tokens (see \S\ref{sec_sim_ops_token}) added in the method appropriate for the model. For demonstration purposes in the original source and target, we \textbf{boldface} and color match areas that <REPHRASE> was applied to, we \textit{italicize} areas that <DEL> was applied to, and \underline{underline} areas that <REORDER> was applied to.
    }
    \label{fig:approach}
\end{figure*}

The following example illustrates the differences and similarities between CS and TS. The sentence ``\textit{Some indigenous groups living in palm-rich areas use palms to make many of their necessary items and food.}'' from the ASSET \citep{alva-manchego-etal-2020-asset} validation set was simplified by one of the annotators as ``\textit{Groups who live in palm-rich areas use palms to make basic items and food.}''. A CS, in this case, could be ``\textit{People who live in areas with a lot of palm trees use the trees for many things. People can eat the dates that grow on palm trees. People can make many things from palm trees, for example, baskets and plates.}''.\footnote{Simplified by the authors with guidance from a professional cognitive simplifier.} This is an example of the common need in CS to explicitly state assumed prior knowledge, and the need to make the text ``closer'' to the reader (``people'' vs. ``groups''). See  \S\ref{sec:sim_ops_def}.


CS and TS appear to be similar tasks, as similar modifications can be applied in both. CS could even be considered a sub-task of TS, with a target audience of people with cognitive disabilities. 
However, there are two main differences between the two, that we believe motivate further investigation into CS as an independent task. 
First, CS is a well-defined procedure with manuals in multiple languages \citep{plainlanguage.gov,Uziel-Karl2011}, while TS has general guidelines 
and, to the best of our knowledge, no common standards. 
Second, the goal of CS is to simplify texts to provide  \emph{cognitive accessibility} \citep{Yalon-Chamovitz2016}.
This goal of CS can also be at odds with TS's more general goal of improving comprehension, such as when simplifying an article for school students vs. for adults with cognitive disabilities at a similar language proficiency.


As a first step, we explore CS and TS in English, and leave exploration of other languages and intra-language comparisons for future work.

There are very few NLP works that tackle CS.
As such, scarce data is available for training potential CS models.
We propose a methodology to address this gap, by introducing an inductive bias to a model trained on TS, in the form of simplification operations.
We propose a set of simplification operations based on CS manuals, and show that adding inductive bias regarding their use improves performance on the ASSET test set, compared to a strong baseline model.

In addition, we present an English parallel corpus aimed at CS, which we use as a test set.\footnote{This dataset, together with all our code, is publicly available under \href{https://creativecommons.org/licenses/by-nc-sa/4.0/}{CC BY-NC-SA 4.0} on \href{https://github.com/eytan-c/CognitiveSimplification}{GitHub} and \href{https://huggingface.co/datasets/eytanc/FestAbilityTranscripts}{huggingface datasets}.}
We show that when fine-tuning models on TS data, our method improves the models' SARI score on the CS dataset, allowing better task adaptation from TS to CS. 
Finally, we compare how the operations are used in the new CS dataset and existing TS corpora, and show that CS differs from TS not only in goal, but also in data statistics.

\section{Cognitive Simplification}
\label{sec:rw_rw}

The field of cognitive accessibility \citep{Yalon-Chamovitz2009A}
is derived from defining accessibility to include the ability to use services, receive information, and participate in activities, in addition to the more commonly accepted physical ability to reach, navigate, and move in a place. 
This definition codified the accessibility measure of simplifying textual information to address the need of people with cognitive disabilities to understand textual information, i.e., Cognitive Simplification.
Subsequent operationalizations of this notion were carried out by \citet{Uziel-Karl2011} and \citet{Yalon-Chamovitz2016}. In particular, they emphasize the need to preserve as much of the meaning of the original text as possible, without rendering it childish or simplistic, while using the same written language as the original text. Although cognitively simplified texts can be easier to read for people with learning disabilities (such as dyslexia), people with learning disabilities are not the main target audience for them.

NLP research into TS for people with cognitive disabilities is relatively scarce.
Most works focus on measuring the effect of cognitively simplified text on the comprehension of people with cognitive disabilities \citep{Rochford2017, rochford_developing_2021} and without them \citep{Djamasbi2016A,Djamasbi2016B}. A different line of work explored how people with different cognition react to texts at different simplification levels \citep{yaneva-etal-2016-corpus}.

Several works \citep{Feng2009,yaneva-etal-2016-corpus} detail parallel corpora of regular and EasyRead documents, documents that are created via the process of CS. 
Although these works provide details regarding linguistic phenomena in their corpora, we were not able to find any of the corpora detailed therein to run evaluations on.
In addition, we were not able to find any recent works that report results on these corpora, using neural techniques for TS.

Although some preliminary works reference the use of contemporary NLP methods for CS to generate simplification examples \citep[e.g.,][]{rochford_developing_2021}, to the best of our knowledge none provide details regarding the model used, model hyperparameter choices, and evaluation methodology. 
As such, we consider our work to be one of the first to tackle CS as a rigorous, distinct NLP task.\footnote{Contemporaneous work by  \citet{Rennes1647431} also addresses TS for people with cognitive disabilities in Swedish.}

Two other tasks that are related to CS, and use contemporary NLP methods, are text2picto \citep{sevens-text2picto1, sevens-text2picto2} and picto2text \citep{sevens-etal-2015-natural}. These are the tasks of converting text to the Sclera\footnote{\url{http://www.sclera.be/}} and Beta\footnote{\url{https://www.betasymbols.com/}} pictogram languages, designed for people with IDD (intellectual or developmental disabilities), and vise versa. While the output of both tasks can improve access to information for people with cognitive disabilities, we believe this task to be distinct from CS and especially TS, that focus on written and spoken language. 

\section{Other Related Work}
\label{sec:rw}
We would like to highlight key points from \citet{Alva-Manchego2020} relevant to our work that relate to training and evaluation datasets and evaluation metrics.

The main datasets used to train and evaluate TS models are WikiLarge \citep{zhang-lapata-2017-sentence} and Newsela \cite{xu-etal-2015-problems}. 
Both corpora contain matching complex-simple document pairs, whose sentences are automatically or manually aligned to create the datasets. 
In WikiLarge, the matching document pairs are taken from English Wikipedia\footnote{\url{https://en.wikipedia.org/}} and Simple English Wikipedia,\footnote{\url{https://simple.wikipedia.org/}} that aims to be more accessible to people with lower English skills, mainly language learners. 
In Newsela,\footnote{\url{https://newsela.com/data/}} the matching document pairs are articles written professionally at four different reading levels, and are originally intended to be used to teach language skills at different school grade levels.

The latest training datasets, and the current de facto standard for TS training, are WikiAuto and NewselaAuto, created by \citet{jiang-etal-2020-neural} by using a neural CRF sentence alignment model. Both are split into training and validation sets. 
To train their neural CRF aligner, \citet{jiang-etal-2020-neural} also compiled two manually aligned datasets, WikiManual and NewselaManual, split into development, train, and test sets.

The two main datasets used for validation and evaluation of TS models are Turkcorpus \citep{xu-etal-2016-optimizing} and ASSET \citep{alva-manchego-etal-2020-asset}. 
Both contain multiple references for each source sentence (8 and 10 respectively). They are crowdsourced and validated professionally. 

The main metric used for evaluating TS models is SARI \citep{xu-etal-2016-optimizing}, which is computed based on three token-level operations: ADD, KEEP, and DELETE. For the full calculation, see \autoref{app:SARI}.


Many previous works in TS also report BLEU \citep{Papineni2002}. However, several works \citep{sulem-etal-2018-bleu,xu-etal-2016-optimizing}, have shown that BLEU scores are not suitable for the evaluation of TS models. 
Nevertheless, BLEU is still reported, and so we also report it for completeness. 

A contemporaneous work \citep{alva-manchego-etal-2021-un} argued for the value of manual evaluation in TS rather than automatic metrics. We defer this exploration for CS for future work.

Recent works have proposed methods to control TS outputs by prepending special tokens to the input of a TS model, in a similar manner to the one explored in this work. 
Such control allows adjusting the model's outputs to different target audiences, and to control what aspects of the simplification process are applied.
ACCESS \cite{martin-etal-2020-controllable}, and MUSS \cite{Martin2020} both use four structural features of the input-output pairs to define what tokens to prepend during training, and at inference they predefine which tokens to use for all inputs. \citet{sheang-saggion-2021-controllable} add a fifth token to this methodology. \citet{scarton-specia-2018-learning} use a combination of tokens to specify the type of simplification to perform and the grade level to which to simplify to. 
Similarly to these works, we also define special tokens to add to the input at training, while at inference we take a different approach (see \S\ref{sec:experiments}).

Other recent work on TS focuses on particular simplification operations \citep{Zhong_Jiang_Xu_Li_2020,srikanth-li-2021-elaborative}, or on combining different operation modules in a joint model \citep{maddela-etal-2021-controllable}.
\citet{srikanth-li-2021-elaborative} define Elaborative Simplification as simplification by adding information to the source text, rather than just removing redundant information. This aligns with some of our proposed simplification operations (Adding Information and Explicitation, see \S\ref{sec:sim_ops_def}). 
Similarly, \citet{Zhong_Jiang_Xu_Li_2020} focus on whole sentence deletion, which aligns with some operations from our proposed list (Deleting information, and Operations on Sentences).
\citet{maddela-etal-2021-controllable} combine a module for sentence deletion and splitting with a paraphrasing module to generate final simplifications. We discuss all three operations in \S\ref{sec:sim_ops_def}.

\section{Our Approach}\label{sec:approach}

To learn how to simplify a text, a model needs to learn \emph{what} types of modifications to apply to the input and \emph{how} to apply each one. 
These modifications can be categorized into {\it operations}. 
Moreover, since TS has multiple large-scale datasets commonly used for training, while there are hardly any such datasets for CS, incorporation of some form of CS-focused inductive bias into a TS-trained model would be useful to allow it to adapt to the CS task.
The inductive bias could also be useful for improving TS on its own, given the similarities between the two tasks (see \S\ref{sec:results} and \S\ref{sec:sim_compare}).

As such, our hypothesis is that a TS-trained model that was trained to be aware of the use of CS simplification operations, will perform better at TS and adapt better to CS than a model that was trained end-to-end. We will now turn to testing this hypothesis empirically.

\begin{table*}[t]
    \centering
    \small
    \begin{tabular}{c|c|l|cccc|c|c}
\textbf{Task} & \textbf{Train} & \textbf{Model} & \textbf{SARI} & \textbf{ADD} & \textbf{KEEP} & \textbf{DELETE} & \textbf{BLEU} & \textbf{\% Ident.} \B\\ \hline \hline
\multirow{10}{*}{\textbf{TS}} & & GEM T5Base & 30.35 & 3.11 & 62.24 & 25.7 & 0.898 & 40.66\% \T\\
& & GEM BART-Base & 32.16 & 3.11 & 62.17 & 31.21 & 0.888 & 38.16\% \B \\ 
\cline{2-9}
 & \multirow{8}{*}{\rotatebox[origin=c]{90}{\textbf{Auto}}} & T5Large & 32.92 & 2.92 & 61.70 & 34.12 & 0.901 & 39.28\% \T\\
  & & T5Large+Classifier$^{\spadesuit,\spadesuit}$ & 36.90 & \textbf{4.73} & 61.10 & 44.87 & 0.855 & 23.68\% \\
 & & T5Base$^*$ & 32.01 & 3.04 & 61.96 & 31.05 & 0.903 & 35.93\% \\
 & & T5Base+Classifier$^{\spadesuit,\spadesuit}$ & 38.13 & 4.55 & 61.20 & 48.65 & 0.860 & 23.68\% \\
 & & BART-Large$^\spadesuit$ & 36.05 & 4.61 & 61.82 & 41.71 & 0.857 & 19.22\% \\
 & & BART-Large+Classifier$^{\dagger,\spadesuit}$ & \textbf{38.76} & \textbf{4.73} & 60.78 & \textbf{50.78} & 0.845 & 11.70\% \\
 & & BART-Base & 32.43 & 3.24 & 61.91 & 32.13 & 0.885 & 33.70\% \\
 & & BART-Base+Classifier$^{\spadesuit,\spadesuit}$ & 37.22 & 3.87 & 61.93 & 45.86 & 0.874 & 25.91\% \B\\
 \hline\hline
 \multirow{10}{*}{\textbf{CS}} & & GEM T5Base & 19.09 & 1.45 & 41.64 & 14.18 & 0.234 & 70.71\% \T\\
& & GEM BART-Base & 21.77 & 2.43 & 42.63 & 20.24 & 0.238 & 64.17\% \B \\
\cline{2-9}
 & \multirow{8}{*}{\rotatebox[origin=c]{90}{\textbf{Auto}}} & T5Large & 20.02 & 1.67 & 41.38 & 17.01 & 0.231 & 68.54\% \T\\ 
 & &T5Large+Classifier$^{*,*}$ & 21.71 & 2.74 & 41.81 & 20.58 & 0.229 & 57.94\% \\ 
 & &T5Base & 20.66 & 2.04 & 41.86 & 18.07 & 0.237 & 68.22\% \\
 & &T5Base+Classifier$^{\spadesuit,\spadesuit}$ & 26.40 & \textbf{3.02} & 42.19 & 34.01 & 0.222 & 46.11\% \\
 & & BART-Large$^\spadesuit$ & 25.12 & 2.97 & 42.91 & 29.46 & 0.231 & 48.91\% \\
 & & BART-Large+Classifier$^{\ ,\spadesuit}$ & \textbf{27.13} & 2.45 & 42.88 & \textbf{36.05} & 0.221 & 44.55\% \\
 & & BART-Base$^*$ & 23.19 & 2.69 & 42.81 & 24.06 & 0.237 & 58.26\% \\
 & & BART-Base+Classifier$^{\ ,\dagger}$ & 24.54 & 2.13 & 42.92 & 28.58 & 0.226 & 55.14\% \B\\
\end{tabular}
    \vspace{5px}
    \caption{
    Results for all models trained on WikiAuto \citep{jiang-etal-2020-neural} and the GEM baseline models \citep{gehrmann-etal-2021-gem}. Metrics include SARI and the percentage of identical generations (\% Ident.). We also report BLEU for completeness (see text). The highest SARI scores for each fine-tuning setting are \textbf{boldfaced}. We tested significance for the overall SARI scores using Wilcoxon Signed-Rank tests \citep{wilcoxon1945} in two settings. First, for each model type and size, we compared the vanilla model and the matching +Classifier model. Second, compared each GEM baseline model with other models of matching types (T5 and BART). We did so for both TS and CS. Scores with $\rho<0.00001$, $\rho<0.001$, and $\rho<0.01$ are marked with $^\spadesuit$, $^\dagger$, and $^*$ respectively. We mark each +Classifier model with two symbols, respectively for each significance test setting. E.g., in CS, BART-Base+Classifier is not significantly better than BART-Base, but has $\rho<0.001$ when testing against GEM BART-Base.
    }
    \label{tab:results1}
    \squeezeup
\end{table*}

\section{Simplification Operations}
\label{sec:sim_ops}

We adapt existing CS manuals \citep{plainlanguage.gov,PlainLanguageandActionNetwork2011,opm.gov,Management2011,hhs.gov,Uziel-Karl2011} into a list of eight main types of simplification operations. 
Seven of these apply to the simplification instance (SI) level, and the final main type applies to a whole document. 
An SI is a set of one or more sentences in regular language (source) aligned to one or more sentences in simplified language (target).\footnote{See \citet{Alva-Manchego2020}, section 2.1.1.} 
Each main type of operation has multiple sub-operations. For full details, see \autoref{app:ops_def}.

Previous work define different lists of simplification operations \citep{Caseli2009, bott-saggion-2011-unsupervised} or focus on word-level operations (KEEP, ADD, DELETE and sometimes also MOVE \citep{dong-etal-2019-editnts}). Our list is based on independent sources (the CS manuals) and focus on intra- and inter-sentence operations applied mainly to a SI.
\S\ref{sec:sim_ops_def} provides theoretical definitions for each operation. \S\ref{sec_sim_ops_token} describes how we integrate operations into a TS model.

\subsection{Definitions}
\label{sec:sim_ops_def}

Below is the list of definitions for the main types of simplification operations.
\begin{enumerate}[topsep=0pt,itemsep=-1ex,partopsep=1ex,parsep=1ex]
    \item \textbf{Proximation}: Reduces ambiguity in the source by making references in the text ``closer'' to the reader, such as converting a 3rd person point of view to 1st person's.
    \item \textbf{Rephrasing}: Modifying the words used in the source such that simpler words and phrases are used in the target instead of complex, ambiguous, and hard to understand ones.
    \item \textbf{Deleting Information}: Removing words and information from the source via summarization or deletion, to reduce the overall information load on the reader.
    \item \textbf{Adding Information}: Adding information to the target of a SI, that did not appear implicitly or explicitly in the source, mainly through generating relevant examples. 
    \item \textbf{Explicitation}: Explicitly stating or explaining implied knowledge and information from the source\footnote{Explicitation is different from Adding Information since the information that appears ``new'' in the target is actually implied to be understood by all readers in the source.}, and explicitly resolving pronouns and co-references in the target.
    \item \textbf{Intra-Sentence Rearrangement}: Reorder the information content and words of a sentence into a logical and easily followed order.
    \item \textbf{Operations on Sentences}: Operations that apply to a whole sentence, including Sentence Splitting and Sentence Reordering. 
    \item \textbf{Document-Level Operations}: Operations that are applied to a document level, including paragraph reordering, and whole paragraph addition/deletion.
\end{enumerate}
In this paper we focus on the first seven operations. 

All the operations described above make texts easier to understand for any reader \citep{plainlanguage.gov,Uziel-Karl2011}. They are especially important for people with cognitive disabilities, as each in their own way reduces the ``mental load'' required from a reader to understand a given text. For example, ``Adding Information'' by providing examples makes general or abstract concepts more concrete to a reader;
``Explicitation'' by clearly stating implied prior knowledge eliminates the need to query that knowledge from memory; and ``Proximation'' by changing passive voice to active voice makes a sentence easier to follow, since ``\textit{Active voice makes it clear who is supposed to do what.}''.\footnote{Federal Plain Language Guide, Section III.a.1., \citep{PlainLanguageandActionNetwork2011}}

\subsection{Special Tokens for Operations}
\label{sec_sim_ops_token}

This section describes a method for introducing inductive bias regarding the use of operations to a TS model.
For each operation, we create a special token that is added to an SI such that the model would learn to predict the token at inference. See \autoref{fig:approach} for an example.
For each operation, we formulate simple rules that can be applied automatically to determine whether it took place in a given SI. These rules depend on the source and target together, and cannot be discerned deterministically based on the source. 
To prevent overlap between operations that share similar indicators, such as Adding Information and Explicitation (when stating implied prior knowledge), we map the first seven operations into 9 unique tokens: Proximation to <\textsc{prox}>; Rephrasing to <\textsc{rephrase}>; Deleting Information to <\textsc{del}>; Adding Information to <\textsc{add}> and <\textsc{example}>; Explicitation to <\textsc{add}>, <\textsc{explain}>, and <\textsc{explicit}>; Intra-sentence Rearrangement to <\textsc{reorder}>; and Operations on Sentences to <\textsc{reorder}> and <\textsc{split}>. For a full description on the rules used to identify each token, see \autoref{app:token_ident}.

While the use of simple rules to assign operation tokens to SIs is noisy, we see its quality as sufficient for testing our main hypothesis, namely about the value of the inductive bias implied by the operations.
We do not stipulate that our operation classification is optimal, and leave the exploration of more sophisticated methods for future work.

To validate our automatic operation token assignment, we asked an in-house human annotator to manually assign operation tokens to 50 random SIs from the WikiAuto training set according to their definition in \S\ref{sec:sim_ops_def}. 
Using these labels as ground truth, our automatic identification rules achieve a micro precision, recall, and F1 scores of 60.3\%, 90.1\%, and 72.2\% respectively. 
The main fall in F-score is the accuracy of the $<$\textsc{add}$>$ operation, which is assigned by an admittedly over-simplistic rule. The two other most frequent operations have F-scores of around 90\%.
For further details, see \autoref{app:token_ident_scores}. 

We further validated the reliability of the annotation by assigning a co-author of this paper to independently complete the same manual annotation task. This resulted in a remarkably high inter-annotator agreement. Indeed, measured by Cohen's $\kappa$, we get an agreement of $\kappa=0.84$ for the <\textsc{rephrase}> operation, and perfect agreement for other operations.
Taken together, these scores indicate the reliability of the automatic token assignment we employ, at least at the aggregate level. 


\section{Simplification Experiments}\label{sec:experiments}

We use the huggingface\footnote{\url{https://huggingface.co/}} API to fine-tune pretrained language models. We select T5 \citep{2020t5} and BART \citep{lewis-etal-2020-bart} model architectures of two sizes each, Base and Large, to align with the recently published GEM benchmark's \citep{gehrmann-etal-2021-gem} official baseline for TS that uses these two model architectures. In addition, we wanted to test if results are consistent across model architectures. 

\subsection{Training Setting}
\label{sec:train_data}

The main dataset we use for fine-tuning is WikiAuto \citep{jiang-etal-2020-neural}, the automatic alignment of WikiLarge \citep{zhang-lapata-2017-sentence}. This dataset contains 483802/20000 SIs for training/validation respectively, and is the standard dataset used in recent works for TS training. This is also the training set used in the GEM benchmark.

We also experiment with a non-standard training setting, using the manually aligned datasets WikiManual and NewselaManual from \citet{jiang-etal-2020-neural}, who used these datasets to train their respective automatic alignment models for the WikiLarge and Newsela corpora. We experiment with this setting since both datasets as well as our new CS dataset are manually aligned, and manual alignments can potentially capture more complex simplification phenomena. This dataset has 11728/1418 SIs in training/validation sets.

\paragraph{Models.} For each model architecture and size, and each dataset, we fine-tune the model on two different settings: baseline and +Classifier. 
In the baseline setting, the model receives as input the source text, and the target output is the correct simplified sentence. This is the standard methodology used to train TS models. 
In the +Classifier setting, our goal is to force the model to predict simplification operations while simplifying the source sentence. For each model architecture this is achieved differently. 
For T5, since you can bind particular masking tokens to particular spans of the input, we format the input and target for the model such that a mask is bound to the operation tokens and the target remains the simplification. 
For BART, since masks cannot be bound to particular spans, we prepend a masking token to the source and prepend the simplification operations to the target. 
We illustrate both methods in \autoref{fig:approach}.

All models are fine-tuned on a single 24GB RAM GPU for 3 epochs, using a constant learning rate of $10^{-4}$ and the Adafactor optimizer \citep{DBLP:journals/corr/abs-1804-04235}.
At inference, we use beam search with 4 beams and early stopping. We do not perform hyperparameter tuning. Due to computational limitations, we train one model of each (architecture, size, type, training data) combination. 

We also compare each model architecture against the respective GEM baseline using a notebook provided by the original authors.

\subsection{Evaluation Datasets}
\label{sec:eval_data}

All models are evaluated on the ASSET \citep{alva-manchego-etal-2020-asset} test set, which contains 359 SIs. 
This is the standard dataset for evaluating TS models, since it provides multiple reference simplifications for each source sentence. 
The way we decided whether a particular operation is applied to a source sentence in ASSET is by majority of the ten references, meaning, we consider an operation taking place only if more than 50\% of annotators in ASSET used it in their simplifications of that source. 
In \autoref{app:sim_ops_details} we provide more details on the counts of actions in each dataset. 

In addition, we evaluate each model on a new Cognitive Simplification test set, called FestAbility Transcripts. 
This dataset contains aligned transcripts of the virtual accessibility conference FestAbility\footnote{\url{https://www.festability.org/}} held in 2020 during the COVID-19 pandemic. 
The conference was simplified live according to the Yalon Method\footnote{\url{https://www.yalonmethod.com/}}, and the transcripts were manually aligned by the authors to create 321 SIs. 
We use this dataset to test each model's performance in adapting from a TS setting to a CS one.
\autoref{tab:FestAbilityDetails} provides some details into the content of this dataset.

\begin{table}[h]
    \centering
    \begin{tabular}{c|c}
         \textbf{Metric} & \textbf{Value} \\
         \hline \hline
        Unique Tokens – source & 1452 \\
        Unique Tokens – target  &  996 \\
        Shared Tokens  &  798 \\
        TER & 0.92 \\
        Token Length Ratio & 0.95 \\
        Nbchars Ratio & 1.14 \\
        Levinstein Similarity & 46.29 \\
        Wordrank Ratio & 0.83 \\
        Deptree Depth Ratio & 1.11 \\
    \end{tabular}
    \caption{Details for the new FestAbility Dataset. Using a SentencePiece tokenizer, we report the number of unique tokens in the source sentences and the target simplifications, and the number of shared tokens between them. We also report the four metrics from \citet{martin-etal-2020-controllable,Martin2020} for future comparisons between FestAbility and other datasets. }
    \label{tab:FestAbilityDetails}
\end{table}

We report SARI\footnote{Using the EASSE \citep{alva-manchego-etal-2019-easse} implementation of the metric.} \cite{xu-etal-2016-optimizing} for each model on each test set, and we also report separately the scores for each token-level operation (ADD, KEEP, DELETE) that are averaged together to compute SARI.
For completeness, we report BLEU scores for each model as well. However, we should note that according to \citet{sulem-etal-2018-bleu} and \citet{alva-manchego-etal-2021-un}, BLEU is not a suitable metric for evaluating text simplification models.
We also report what percentage of test outputs are identical to the source for each model.

\begin{figure*}[tbhp]
\begin{subfigure}{\columnwidth}
    \centering\includegraphics[width=\columnwidth]{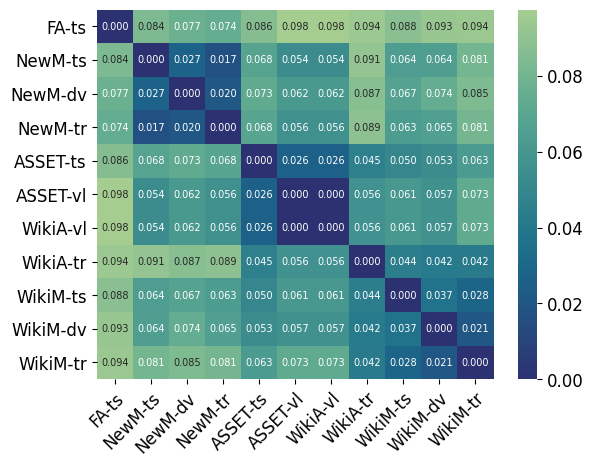}
    \caption{$\overline{JSD}$ distances between distributions
    }
    \label{fig:JSD}
\end{subfigure}
\hspace{10pt}
\begin{subfigure}{\columnwidth}
    \centering\includegraphics[width=\columnwidth]{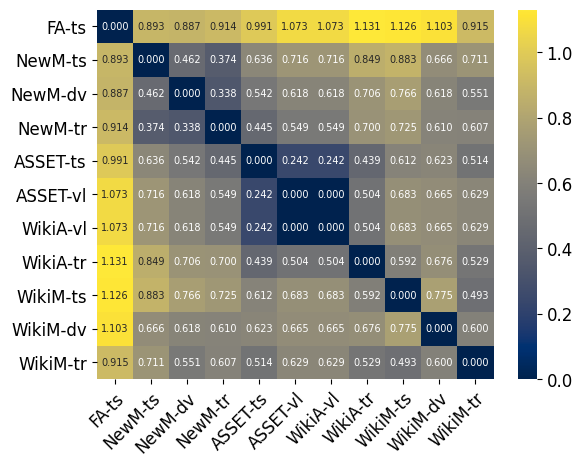}
    \caption{$\ell_2$ distances between  correlation matrices
    }
    \label{fig:l_2}
\end{subfigure} 
    \caption{
    Heatmaps of the distances between dataset sub-sets. We shorten dataset names as follows: FA=FestAbility, NewM=NewselaManual, WikiM/A=WikiManual/Auto. The final two letters signify ts=test, vl=valid, dv=dev, and tr=train sets.
    For each sub-set pair, we report the numerical distance in the matching cell.
    }
    \label{fig:distances}
    \squeezeup
\end{figure*}

\section{Results}
\label{sec:results}

Our main results are presented in \autoref{tab:results1}. Results on TS show that when trained on the standard WikiAuto dataset, the +Classifier variant of a model outperforms the baseline's SARI score in all cases, with 3.98 points for T5Large, 6.12 points for T5Base, 2.71 points for BART-Large, and 4.79 points for BART-Base. These are substantial improvements, considerably larger than differences in SARI scores between model sizes of the same variant, except for the BART baseline models. The difference between the T5 baseline models is 0.91 points, T5+Classifier models is 1.23, the BART baseline is 3.62 points, and the BART+Classifier models is 1.54 points.

Focusing on CS performance, we find that the +Classifier variants achieved superior results for all model architectures and sizes. 
The improvement differs by architecture and size, with the largest difference being of 5.74 SARI point for the T5Base models trained on WikiAuto. The best performance is again obtained by the BART-Large+Classifier model, and is at least 2.01 SARI points higher than the score obtained by any baseline variant.

With respect to the Manual dataset training setting, we see similar trends. 
In particular, the +Classifier models outperform baseline models, and the best performing model is still BART-Large+Classifier. Due to space limitations, we discuss the results on this dataset in \autoref{app:manual}.

Taken together, our results demonstrate the effectiveness of incorporating inductive bias using simplification operations for both TS and CS.

In order to ensure that the experimental setup we use is comparable in performance with the standard practice in the field of TS, we experiment with the original GEM baseline code-base, and our hyperparameter settings were chosen according to it. The results of models trained according to this code-base are indeed comparable to models of matching sizes of the baseline variants.

We further validated our results with significance tests, following the guidelines of \citet{dror-etal-2018-hitchhikers}. 
We used the Wilcoxon Signed-Ranked \citep{wilcoxon1945} test as our main significance test. 
We compared each vanilla and +Classifier model pair, and also each model of a particular type (T5 and BART) to their respective GEM baselines. 
The results are shown in \autoref{tab:results1}. 
Almost all tests, with only six exceptions, are significant with at least $\rho<0.01$ and most with $\rho<0.00001$. These results further support the validity of our analysis.

We attribute the improved performance of all +Classifier models to improvements in the token-level operations scores for ADD and DELETE. In the standard training setting on WikiAuto, all +Classifier models achieve substantially higher ADD and DELETE scores than their same-sized baseline counterparts, while all models achieve similar KEEP scores. Interestingly, for the BART models, the difference in ADD scores is less substantial than for the T5 models.

\section{
Simplification Dataset Comparison}
\label{sec:sim_compare}

We compare simplification datasets with respect to how the simplification operations are used in each. We show that simplification operations can also be used to better characterize such datasets.

We analyze all available sub-sets (development, train, validation, and test) of all datasets, to provide a fine-grained analysis.
We consider test sub-sets of datasets, to better understand the results of \S\ref{sec:results}. 
This analysis was done after-the-fact, and did not influence the development of the models.\footnote{We analyze the test sets also because the CS dataset only contains a test set at this point, due to their small size.} 

The results presented in this section show that CS is different from TS in how the operations are applied. They also surface the known relationships between the datasets, validating our analysis. We believe that this type of aggregate analysis can be confidently performed given the validation at the end of \S\ref{sec_sim_ops_token}, but acknowledge that the token assignment is noisy. 

To understand how each simplification operation is applied individually, we compute the frequency with which each operation is applied in a given sub-set. These frequencies can be viewed as defining random variables $X_o^S$, stating the probability that each simplification operation $o$ is used in a particular SI in sub-set $S$. As such, to understand the distance between sub-sets with respect to the individual application of each operation, we can compute the mean Jensen-Shannon distance \citep{JSD1,JSD2} (which we mark $\overline{JSD}$) between matching random variables in different sub-sets. For further details on the action distributions for each dataset, see \autoref{app:sim_ops_details}.

As can be seen in \autoref{fig:JSD}, all sub-sets have $\overline{JSD} < 0.1$ from one another, which is not a large distance. 
However, we are still able to see distinct clusters for each dataset, with subsets having $\overline{JSD}<0.04$ within clusters and $\overline{JSD}>0.04$ to other sub-sets.\footnote{For reference, if $p=(0.557,0.443)$ and $q=(0.5,0.5)$, then $JSD(p,q)=0.0403$.}  
Interestingly, WikiAuto-test is closer to the WikiManual cluster than it is to WikiAuto-valid, which could be explained by the fact that WikiAuto was created based on the matching of complex-simple sentences presented in WikiManual. 
In addition, WikiAuto-valid and ASSET-valid appear to be identical, which could be explained by the fact that the source for ASSET-valid was taken from WikiAuto-valid.
Regarding the CS dataset FestAbility, it is $\overline{JSD}>0.07$ from all other sub-sets, and is the farthest sub-set from WikiAuto, ASSET, and WikiManual clusters, and the second or third farthest from sub-sets in the NewselaManual cluster.

To understand how simplification operation are applied together, we computed the Pearson correlations of the co-occurrence of each operation pair in a given subset $S$, to create a correlation matrix $M^S$. We then computed the pair-wise $\ell_2$-distance between matrices. Results are in \autoref{fig:l_2}.

As can be seen in \autoref{fig:l_2}, the clusters of closest sub-sets are maintained for NewselaManual, and for ASSET and WikiAuto-val 
, while the sub-sets of WikiManual are no longer closest to one another. Also, WikiAuto-train is similarly distant from both WikiAuto-val and the WikiManual sub-sets, unlike when comparing with $\overline{JSD}$. In this setting, the FestAbility dataset is the most distant sub-set from all other sub-sets, with $d_{\ell_2}>0.88$ from all of them. All other sub-sets are $d_{\ell_2}<0.75$ from one another, except NewselaManual-test from WikiAuto-train and WikiManual-test with $d_{\ell_2}=0.85$ and $d_{\ell_2}=0.88$ respectively.

Taken together, these results show that while each individual operation is applied with similar probability in every dataset, the operations are applied together differently. In CS in particular, they are applied in a more distinct fashion than in TS. 

The difference in operation application could be attributed to the different domains from which each dataset pulls its sentences. In our CS dataset, all sentences are transcripts of human speech, taken from a formal conference. Thus, they may contain more informal language than a Wikipedia article. Given our datasets, we therefore cannot differentiate between domain difference and task difference. However, we are currently compiling a larger dataset for CS that contains more formal language, that will enable such analysis.


The analysis here can provide additional insight as to the performance patterns of the different models (\S\ref{sec:results}). Since each operation is applied individually under a similar distribution in TS and CS, the +Classifier models could have potentially learned indicators of when to apply each action individually when training on TS. 
This could have been useful when adapting to CS, especially given that the operations co-occur differently in TS and CS.

\section{Conclusion and Future Work}

We formulated the task of Cognitive Simplification as an NLP task, and discussed its similarities and dissimilarities from the well-researched task of TS. The two tasks are similar in the types of simplification operations that are applied in each, and different in the distribution in which the operations are applied. They also differ in their target audience, at least when using standard datasets.
We further release with this paper a readily available dataset directed at CS, providing a test set to evaluate CS models on.

Attempting to overcome the absence of training data for CS, we showed that by introducing to a TS-trained model inductive bias as to the  simplification operations that need to be performed on the input, the model is able to better adapt to CS. We also showed that TS-trained models that are trained to predict simplification operations perform better than their baseline counterparts on TS.

We believe that comparing how simplification operations are applied in different languages can provide valuable insights into understanding the task of Text Simplification better. Future work will further explore the relation between the distribution of operations and the ability of the model to generalize to different domains and task formulations. Such an inquiry may reveal that simplification operations provide not only inductive bias, but also an analytical tool for comparing datasets and variants of TS.  There are TS datasets in many languages, including Swedish \citep{rennes-jonsson-2015-tool}, Spanish \citep{Saggion2015}, German \citep{sauberli-etal-2020-benchmarking,battisti-etal-2020-corpus}, Danish \citep{klerke-sogaard-2012-dsim}, Portuguese \citep{leal-etal-2018-nontrivial}, and Russian \citep{dmitrieva-tiedemann-2021-creating}. We plan to compare these datasets in terms of their distribution of operations, so as to empirically characterize whether the notion of text simplification implicit in these datasets is similar or not.

We hope that our findings will spark interest in CS, as there is much more to solve in creating automatic simplification systems for people with cognitive disabilities. As stated above, we are currently working on compiling a larger and more robust CS dataset, that will enable improvements in CS technology, and allow to tease apart domain effects in the differences between TS and CS from more fundamental differences between the tasks.

\section*{Ethical Considerations}
\paragraph{Use of existing datasets.}
The WikiAuto, WikiManual \citep{jiang-etal-2020-neural}, and ASSET \citep{alva-manchego-etal-2020-asset} datasets are publicly available. We took the WikiAuto and ASSET from the huggingface dataset hub,\footnote{\url{https://huggingface.co/docs/datasets/}} and WikiManual from the authors' GitHub.\footnote{\url{https://github.com/chaojiang06/wiki-auto}} We used and received access to Newsela with accordance to Newsela's terms of service. 

\paragraph{The released FestAbility dataset.} The FestAbility conference is available for viewing online, and we received approval to redistribute the simplifications and transcripts from the organization that simplified the conference.\footnote{\url{https://www.yalonmethod.com/}} The text in these transcripts deals with the following subjects: rights of people with cognitive disabilities, arts and performing arts in particular, accessibility, and personal stories. None of the text is offensive or discriminatory in any way. Free public access to this dataset is available for future research under \href{https://creativecommons.org/licenses/by-nc-sa/4.0/}{CC BY-NC-SA 4.0} on GitHub at \url{https:/github.com/eytan-c/CognitiveSimplification} and as a huggingface dataset at \url{https://huggingface.co/datasets/eytanc/FestAbilityTranscripts}.

\paragraph{Ethical risks.}
We do not see any immediate adverse effects that our methodology and dataset can lead to. On the contrary, further research into CS from an NLP context can only provide benefits to people with cognitive disabilities.

\paragraph{Other Considerations}
\citet{gooding-2022-ethical} recently presented multiple different ethical considerations for text simplification research. These include stating explicitly the target audience for TS, using appropriate datasets, and evaluating using appropriate measures, among others. While contemporaneous, our paper aligns with the claims that \citet{gooding-2022-ethical} present with how we define the task of CS. Furthermore, the methodology presented in \S\ref{sec:sim_compare} can be used to empirically measure some of the risks presented in Section 3 of \citet{gooding-2022-ethical}. 

\section*{Limitations}
\paragraph{Computational limitations.}
Each model trained in \S\ref{sec:experiments} requires a long time to train on the largest GPU available to the authors, with the largest models taking several days to complete the training. See \autoref{app:train_time} for details. These resources therefore prohibit experimentation with larger models.
\paragraph{Comparison to other TS systems.}
The TS literature contains many TS systems, using many different techniques (such as \citet{martin-etal-2020-controllable,Martin2020,sheang-saggion-2021-controllable,scarton-specia-2018-learning,Zhong_Jiang_Xu_Li_2020,maddela-etal-2021-controllable,zhao-etal-2018-integrating, zhang-lapata-2017-sentence}). Any one of these systems could be used as well for CS, and such a comparison is warranted. The goal of this paper however is to highlight the need and possibilities of further research into CS, and provide initial benchmarks and tools to do so. We do not presume that our methodology of adding simplification operations is the best methodology for CS. We leave investigating the answer to this question for future research. The authors are currently working on answering this question, in particular in conjunction with releasing additional CS data. 
\paragraph{Using additional datasets.}
Although we did get permission to use NewselaAuto as a training dataset, we did not train models with that dataset to report results on. The reasoning behind this decision that we wanted the main results of this paper to be easily reproducible, and while WikiAuto is readily available for use by all, access to Newsela is provided under a restrictive license.
\paragraph{Adding simplification operations.}
The methodology proposed in the paper to add simplification operations to SI uses simplistic rules to do so. Some of the operations can be quite difficult to identify, even for humans. We believe that there probably is a better methodology for identifying the simplification operations, and leave identifying such a methodology for future research.

\section*{Acknowledgements}

This work was partially supported by the Israel Science Foundation (grant No. 929/17). We would like to thank Prof. Shira Yalon-Chamovitz for helpful discussions and for providing us with the cognitive simplification guidelines and data. We would like to thank the authors of the GEM baseline for simplification, for providing the source code used to train their baseline models. We would also like to acknowledge the Adi Lautman Interdisciplinary Program, for providing the fertile ground in which the initial idea for this project grew.

\bibliography{anthology,custom}
\bibliographystyle{acl_natbib}

\appendix
\section{Simplification Operation Definitions}
\label{app:ops_def}
In this section we describe in more detail the different simplification operations, providing full details for each, including sub-operations.

This list is based on cognitive simplification manuals, and includes 2 levels of operations, as many particular operations share similar goals. We describe the similar goals as ``Main Operations'', and this is the list provided in the main paper. In here, we describe in detail all sub-operations as well.

As explained in the main paper, we focus mainly on the operations that are performed on simplification instances (SIs). 
We do so both to align with existing research of TS, and to conform with how the simplification manuals describe the process of CS. 
In addition, we also describe ``Document Level'' operations.
These ``Document Level'' operations are not distinct to CS, but have an important role in that task.

For each operation, we also describe what type of modification to the source of a SI is this operation aimed at: a modification of its syntactic structure or the modification of its lexical content (i.e., the words used in the SI). 
We deem the former a structural modification, and the latter a lexical modification.  
Some operations perform both, but in such cases we chose to assign the type of modification that subsumes the other. 
For example, Sentence Splitting is a structural modification, since it aims to modify the structure of the original text by splitting a sentence into two or more sentences in the simplification. 
This structural change might require changing words used in the target (i.e., a lexical modification) but those changes are part of the structural modification.

\begin{enumerate}
    \item \textbf{Proximation}: Proximation is the process of making references in the text closer 
    to the reader, meaning explicit and more relatable. 
    This can be by changing the point of view of the sentence from 3rd to 2nd and/or 1st person, by changing the tenses of verbs to easier to understand tenses\footnote{For example, Present Tenses are generally easier to comprehend than Future Tenses. Another example: in English, Perfect Tenses harder to understand and should usually be converted to other tenses.}, or by converting Passive voiced sentences to Active voiced ones\footnote{Multiple CS manuals state that sentences with an active voice are easier to understand than sentences in passive voice \cite{PlainLanguageandActionNetwork2011,Uziel-Karl2011}. From the Federal Plain Language Guide, Section III.a.1.i, page 20: ``\textit{Active voice makes it clear who is supposed to do what. It eliminates ambiguity about responsibilities. Not ``It must be done.'', but ``You must do it.''. Passive voice obscures who is responsible for what ...}''. \citet{Uziel-Karl2011} even explicitly state that every passive voiced sentence needs to be converted to active voice.}. 
    This reduces the potential ambiguity in the source and makes the target more personal, and thus more easily understood to people with cognitive disabilities.
    
    Proximation, and all of its sub operations, are \textit{structural} modifications, since their goal is to transform the syntax of the sentence (tense, voice, etc.).
    
    \item \textbf{Rephrasing}: Modifying the words used in the source such that simpler words and phrases are used in the target instead of complex, ambiguous, and hard to understand ones. Simpler words and simpler phrases makes the text easier to understand for people with lower language comprehension skills, such as those with cognitive disabilities. A rephrasing can be finding a simple synonym to a complex word, but also converting words to phrases and vise-versa.
    Since Rephrasing changes the words used in a sentence, it is a lexical modification.
    \item \textbf{Deleting Information}: A main part of simplifying a text is deciding which information is irrelevant or surplus to a reader's comprehension, and removing it from the text. By lowering the information load on the reader, his or her ability to comprehend the text increases. Deleting Information comes in two main types, Removal and Summarization. We chose to assign both into Deleting Information, since in both some of the \textit{information content}\footnote{See \autoref{app:info_content} for a discussion on this topic} of the source is lost in the target, either directly (Removal) or indirectly (Summarization). 
    
    Deleting Information is a \textit{lexical} modification. 
    
    \item \textbf{Adding Information}: This operation includes adding information to the simplification that never appeared in the source. It includes only one sub-operation, Example Generation, since this is the only type of novel information that can appear in the target of an SI. Any other apparent ``new information'' is usually implicit information that is part of the source, and requires Explicitation in the target. 
    
    However, finding precise distinctions between new information in the target that is 100\% new and new information in the target that is implicit information from the source is a difficult task. As such, we chose to have a general ``Adding Information'' operation for exactly the type of new information in the target that cannot be precisely associated either as an Explicitation or Example Generation.
    
    Adding information is a \textit{lexical} modification.
    \item \textbf{Explicitation}: 
    Many of the texts we read contain implicit information that the writer assumes the reader has prior knowledge of. 
    During simplification, this implicit information will need an explanation or elaboration upon, so that the reader can understand the text. 
    
    This could be achieved by Explanation Generation: explaining the meaning of particular terms and phrases, or explicitly stating the logic and reasoning behind a particular passage in the text. These explanations are crucial for people with cognitive disabilities to understand texts, since they sometimes lack prior common knowledge in many domains. 
    
    We consider both Explanation Generation and Example Generation (from the previous main operation) to be forms of \textit{Elaborative Simplification} \cite{srikanth-li-2021-elaborative}. We create a distinction between the two to differentiate between ``new information'' in the simplification that is from the implicit information of the source and ``new information'' from the potentially relevant information of the source. See  \autoref{app:info_content}.
    
    In addition, the source might contain pronouns that the writer assumes their co-references can be resolved easily from the text. However, in most cases, people with cognitive disabilities would not necessarily be able to resolve pronoun co-references. As such, most pronouns should be converted in the target to their explicit references. This is Pronoun Explicitation.
    
    Both types of Explicitation are \textit{lexical} modifications.
    \item \textbf{Intra-Sentence Rearrangement}: At times, the clauses of a sentence can be ordered in such a way that make it harder to comprehend due to its clauses being out of the ``correct'' logical order. In addition, for many reasons, the ordering of the subject, verb, and object can be out of the ``correct'' order. When information is presented out of order, it makes the text harder to comprehend, especially for people with cognitive disabilities. Semantic Rearrangement is presenting the information content of a sentence in the source of an SI in a logical and easily followed order, and is a \textit{structural} modification. \footnote{This passage is written on purpose in a convoluted order, to demonstrate to the reader the importance of order to text comprehension.}
    \item \textbf{Operations on Sentences}: There are often simplification operations that are applied on a whole sentence that is part of a SI, rather than applying to an internal part of a sentence. This includes Sentence Splitting, and also Sentence Reordering. 
    
    Splitting long sentences into shorter ones makes texts easier to comprehend by reducing the information load of each sentence. Rearranging the sentences of a paragraph into a correct logical/temporal order also makes a text easier to comprehend, for the same reasons explained above in Intra-Sentence Rearrangement. 
    
    Sentence Operations are \textit{structural} modifications.
    
    \item \textbf{Document Level Operations}\footnote{As stated in the main paper, we focus mainly on the SI operations, and less on the document level operations. We still state them here to present a complete picture.}: In some cases, when simplifying long texts organized as documents and/or documents with subsections, more overarching operations need to be applied. These are almost always modification of \textit{structure}, since information needs to be ordered correctly, as explained in the previous two Main Action types. This can include full chapter/sub-document reordering and full paragraph reordering, but can also cross paragraph reordering of sentences and paragraph splitting. 
    
    In addition, there are \textit{lexical} modifications that we consider a Document Level Operations. These are Adding Paragraphs and Adding Chapters that didn't exist in the original document, and Deleting Paragraphs and Deleting Chapters from the original document. The additions of paragraphs or chapters usually explain particular concepts or ideas crucial to comprehending the document, while deleting paragraphs or chapters in their entirety is usually because the information they provide is not crucial for comprehending the main idea of the document. 
    
\end{enumerate}

    \begin{figure*}[tbhp]
\begin{subfigure}{\columnwidth}
    \centering\includegraphics[width=0.8\columnwidth]{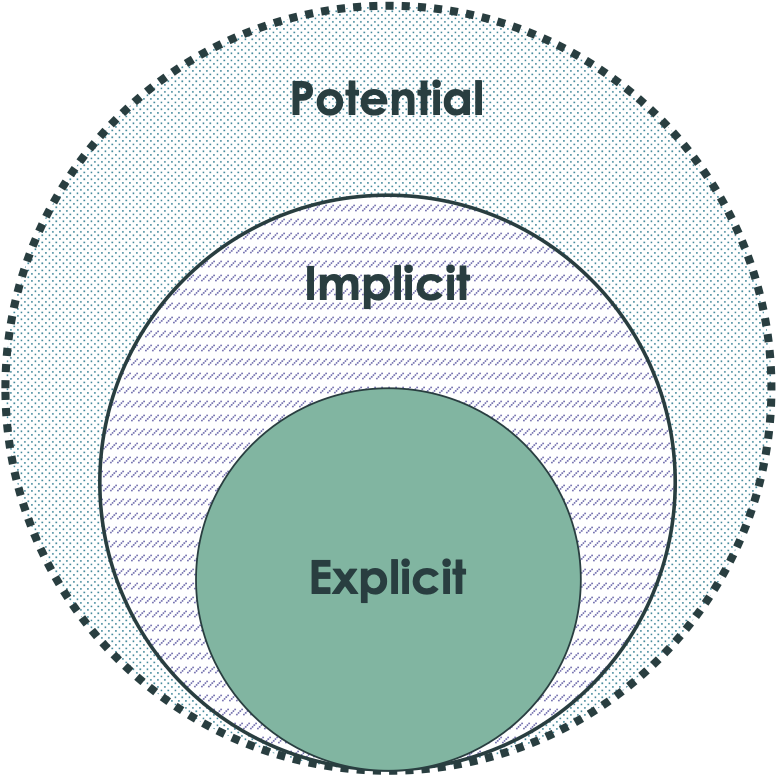}
    \caption{Source\label{fig:source_info}
    }
\end{subfigure}
\begin{subfigure}{\columnwidth}
    \centering\includegraphics[width=0.8\columnwidth]{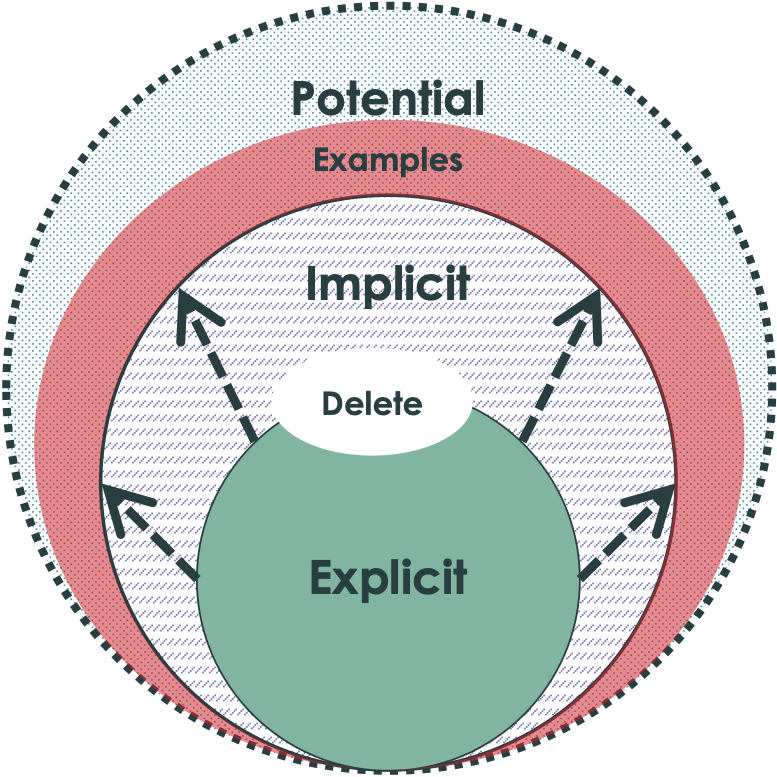}
    \caption{Target\label{fig:target_info}
    }
\end{subfigure} 
    \caption{
    Diagrams showing the transformation of \textit{Information Content} between the Source and Target in a simplification instance.
    }
    \label{fig:info_content}
\end{figure*}

\subsection{Modifying the Information Content of Simplification Instances}
\label{app:info_content}

We would like to propose a clear definition of how the information content of a text is modified during the process of simplification. For this, we define the \textit{explicit} information content of a text as being the information that is encoded by the exact words of the text. 
Each text, in addition to the information explicitly stated by words used in the text, also encodes \textit{implicit} information about those words and the subjects they describe. This includes assumed prior knowledge related to the subject of the text or the use of phrases in it, references to other parts of the text, understanding the logic and reasoning behind the information described in the text, and more. 
The \textit{potentially relevant} information can be defined as all the potential utterances that describe information and knowledge that can be relevant to a particular text. 
This information is not explicitly stated in the source or needs to be implied to understand it, and if the information appears in the target, the decision to include the particular utterance can't be uniquely predicted given the source.
In essence, \textit{potentially relevant} is net new information that can appear in the target. For CS, this happens mainly in the form of Example Generations, and the particular example chosen for a given simplification could easily be switched with other examples.

Using these three types of information content, we can better define the process of CS, and the distinctions between the simplification operations of \textbf{Adding Information}, \textbf{Explicitation}, and \textbf{Deleting Information}.

We can formulate the process of CS as minimizing the distance between the explicit and implicit information content of a text \textit{as much as possible}, while removing redundant or surplus information and adding relevant novel examples, all in the goal of making the text more comprehensible to people with cognitive disabilities. This is juxtaposed with TS, in which the distance between explicit and implicit information content is minimized, but not to the maximal degree.

\section{Special Token Identification}
\label{app:token_ident}
Each of the operations described in \autoref{app:ops_def} can be potentially identified using multiple different methods. 
In this appendix we describe how we identified each operation and sub-operation in order to prepend the relevant special token as seen in \autoref{fig:approach}. 

For the scope of this work, we chose to use deterministic heuristics that can be applied automatically. Although they create noisy classifications, we chose the heuristics such that hey have an emphasis on Precision rather than Recall, and so we find them sufficient for our work.

Most of the operations below are analyzed in the context of simplification instances, and we describe in input as the ``source'' and simplification as the ``target''. These will be mathematically noted as $S$ and $T$ respectively when relevant.

The full code that we used to identify these operations is available on \href{https:/github.com/eytan-c/CognitiveSimplification}{GitHub}.

\begin{enumerate}
    \item \textbf{Proximation}: 
    All of these operations are tested on a word by word basis using the Universal Dependency parse trees of the source and the target.
        \begin{enumerate}
            \item \textit{Change of person point of view}: 
                We check if there was a change in person POV from 3rd to 2nd, 3rd to 1st, or 2nd to 1st.
            \item \textit{Modify verb tense}: We check if the verbs in the target are in a different tense than the matching verbs in the source. 
            \item \textit{Passive-Active Substitution}: We check if there exist any passive verbs in the source that share meaning with active verbs in the target.
        \end{enumerate}
    Any SI that has a Proximation operations was prepended with the token $<$\textsc{prox}$>$.
    \item \textbf{Rephrasing}: 
        A rephrasing operation will follow the format of replacing one or more words from the source with one or more words with similar meaning in the target. Thus, to identify a rephrasing, we tested every word in the source sentence that did not appear in the target against known paraphrase databases for the relevant language (such as SPPDB \cite{pavlick-callison-burch-2016-simple} for English) to see if one of their relevant paraphrases appears in the target.
        
        Phrasing this mathematically, for every word $w\in S\setminus T$, we check if $pp(w)\subset T$, where $pp(w)$ is the result of applying a rule from a paraphrase database on $w$.
        
        \begin{enumerate}
            \item \textit{Simple synonym}: These operations are defined when one word is paraphrased to another single word.
            \item \textit{Paraphrasing}
                \begin{enumerate}
                    \item \textit{Word-to-Phrase}: Similar to simple synonym, only a single word is paraphrased into a series of words.
                    \item \textit{Phrase-to-Word}: A phrase is converted to a single word. This is discovered by checking all possible combinations of consecutive words in the source that did not appear in the target for possible paraphrases.
                    \item \textit{Phrase-to-Phrase}: Similar to Phrase-to-Word, when the paraphrase rule is to another phrase instead of a single word. 
                \end{enumerate}
        \end{enumerate}
        Any SI that has a Rephrasing operation was prepended with the token $<$\textsc{rephrase}$>$.

\begin{table*}[t]
    \centering
    \small
    \begin{tabular}{c|c|l|cccc|c|c}
\textbf{Task} & \textbf{Train} & \textbf{Model} & \textbf{SARI} & \textbf{ADD} & \textbf{KEEP} & \textbf{DELETE} & \textbf{BLEU} & \textbf{\% Ident.} \B\\ \hline \hline
\multirow{8}{*}{\textbf{TS}} & \multirow{8}{*}{\rotatebox[origin=c]{90}{\textbf{Manual}}} & T5Large & 33.03 & 2.41 & 61.78 & 34.91 & 0.916 & 48.75\% \T\\ 
 & & T5Large+Classifier & 31.78 & 2.34 & 61.27 & 31.71 & 0.909 & 57.66\% \\
 & & T5Base & 30.41 & 1.77 & 62.03 & 27.42 & 0.920 & 56.55\% \\
 & & T5Base+Classifier & 30.48 & 1.87 & 62.35 & 27.21 & 0.920 & 62.12\% \\
 & & BART-Large & 32.27 & 2.85 & 61.27 & 32.69 & 0.888 & 55.43\% \\
 & & BART-Large+Classifier & \textbf{37.66} & \textbf{3.87} & 59.93 & \textbf{49.19} & 0.842 & 31.75\% \\
 & & BART-Base & 31.97 & 1.76 & 61.83 & 32.31 & 0.914 & 55.15\% \\
 & & BART-Base+Classifier & 32.65 & 2.45 & 61.63 & 33.87 & 0.876 & 54.31\% \\
 \hline\hline
 \multirow{8}{*}{\textbf{CS}} & \multirow{8}{*}{\rotatebox[origin=c]{90}{\textbf{Manual}}} &T5Large & 21.10 & 1.43 & 41.98 & 19.91 & 0.234 & 69.16\% \T\\
 & &T5Large+Classifier & 22.43 & 1.21 & 42.78 & 23.30 & 0.235 & 72.27\% \\
 & &T5Base & 20.14 & 1.69 & 42.35 & 16.38 & 0.243 & 72.90\% \\
 & &T5Base+Classifier & 20.21 & 0.89 & 42.26 & 17.47 & 0.239 & 78.82\% \\ 
 & & BART-Large & 21.42 & 1.61 & 42.28 & 20.39 & 0.238 & 75.08\% \\
 & & BART-Large+Classifier & \textbf{26.47} & \textbf{2.34} & 42.13 & \textbf{34.94} & 0.219 & 60.12\% \\
 & & BART-Base & 23.48 & 2.27 & 42.88 & 25.29 & 0.24 & 72.27\% \\
 & & BART-Base+Classifier & 24.22 & 2.26 & 43.52 & 26.87 & 0.242 & 73.52\% \\
\end{tabular}
    \vspace{5px}
    \caption{
    Results for all models fine-tuned on the Manual dataset (see \autoref{app:manual}). Metrics include SARI, the percentage of identical generations (\% Identical). We also report BLEU for completeness (see text). Highest SARI scores for each fine-tuning setting are \textbf{boldfaced}. 
    }
    \label{tab:results2}
    \squeezeup
\end{table*}
    \item \textbf{Deleting Information}: Any words in the source that doesn't appear in the target designate a Deleting Information operation. We discern between Removal and Summarization mainly according to the alignment type. Precisely discerning between the two operations for other alignments types is a more complicated task that cannot be resolved by a simple heuristic, and as such we leave it for future research. For our analysis' purpose, whenever the token length ratio \citep{martin-etal-2020-controllable} between source and target was greater 1.2 than ($|S|/|T| >= 1.2$), or that the percentage of deleted words from the source (i.e., that were removed in the target and were not part of another operation such as Rephrasing) was higher than 30\% and the token length ratio was $>$ 1, we classified it as a Deleting Information operation.
        \begin{enumerate}
            \item \textit{Removal}: If the sentence alignment type of the SI is $M$-to-$0$, we count the operation as \textit{Removal}.
            \item \textit{Summarization}: If the sentence alignment type is $M$-to-$1$, we count the operation as \textit{Summarization}.
        \end{enumerate}
    Any SI that has a Deleting Information operation was prepended with the token $<$\textsc{del}$>$.
    \item \textbf{Adding Information}: To discover if an action was of Adding Information, we check if there are new words in the target, that aren't part of another modification (such as Rephrasing or Passive-Active Substitution) or are function words. Once such words exists, we assume that there is additional explicit information in the target that did not appear in the source. We then test if it is Example Generation or Explanation Generation (see below), and if it is neither, similar to the general classification in Deleting information, if the token length rations between source and target is $<$ 1 (target is longer), we classify as Adding Information.
        \begin{enumerate}
            \item \textit{Example Generation}: If the new words are part of a clause that starts with indicative phrases for providing examples (such as ``e.g.'', ``for example'', ``such as'', and more) we classify this operation as \textit{Example Generation}. This is the only case where we would prepend the SI with the token $<$\textsc{example}$>$.
        \end{enumerate}
    Any SI that satisfied the token length ratio $<$ 1 was prepended with the token $<$\textsc{add}$>$.
    \item \textbf{Explicitation}: 
        From a modeling perspective, we grouped Pronoun Explicitation and Explanation Generation together, since their purpose is similar -- reducing ambiguity in the source that is related to the implicit information and assumptions. However, from a classification perspective, each is discovered differently.
        \begin{enumerate}
            \item \textit{Pronoun Explicitation}: We use a co-reference resolution (CRR) model (Coreferee from Spacy\footnote{\url{https://spacy.io/universe/project/coreferee}}), applied to the concatenated source and target. If the CRR model finds explicit references in the target to pronouns in the source, we classify as Pronoun Explicitation. This is the only case where we would prepend the SI with the token $<$\textsc{explicit}$>$.
            \item \textit{Explanation Generation}: We identify this operation together with Adding Information, since heuristically they can appear very similar. If new words in the target aren't tied to an example, or are tied to a noun phrase in the source that is part of one or more sentences in the target, we assume that this is a form of Explanation. Discerning between the different types of explanation generations is a task for future research, but we list them here for indexing purposes.
            \begin{enumerate}
                \item For term/phrase
                \item For logic/reasoning
                \item For background information
            \end{enumerate}
            Any SI that was identified containing Explanation Generation $<$\textsc{explain}$>$.
        \end{enumerate}
        
    \item \textbf{Intra-Sentence Rearrangement}: This operation is identified when the information order in a text is changed. We use the Universal Dependency parse trees of the source and target to discover rearrangements.
        \begin{enumerate}
            \item \textit{Clause Reordering}: If the clauses in the target appear in a different order than in the source, then this is a \textit{Clause Reordering} operation.
            \item \textit{SVO} Reordering: For each sentence in the source, we check if the order of subject, verb, and object are maintained in  the target. If not, then this is an \textit{SVO Reordering}.
        \end{enumerate}
    Any SI that has an Intra-Sentence Rearrangement operation was prepended with the token $<$\textsc{reorder}$>$.
    \item \textbf{Operations on Sentences}: These operations are checked on a sub-document level, as compared to a simplification instance level.
        \begin{enumerate}
            \item \textit{Sentence Splitting}: This operation is assumed to appear by default in SIs with sentence alignment type of 1-to-$N$. Any such SI was prepended with the $<$\textsc{split}$>$ token.
            \item \textit{Sentence Rearrangement}: Part of the manual alignment process, the original ordering of sentences in the source sub-document and be compared to the order of the original sentences according to their alignment to the target sub-document. So, if the source sub-document consists of sentence $[s_1,s_2,s_3,...,s_n]$ and their alignment to the target sub-document sentences is some permutation of their indexes $I$, such that the source sentences ordered by the target's order is $[s_{i_1}, s_{i_2}, ..., s_{i_n}]$, we look for the longest increase sub-sequence in this permutation $L\subset I$. Any sentence indexed by $i_j\notin L$ is a Sentence Rearrangement. 
            From an SI perspective, a similar analysis was done for \textit{Clause Reordering}, in order to discover to which SIs to prepend the $<$\textsc{reorder}$>$ token.
        \end{enumerate}
    \item \textbf{Document Level Operations}: We list here the Document Level Operations, but for our analysis we only focused on identifying Adding/Deleting Paragraphs and Sub-documents, which were respectively classified as Adding/Deleting Information. In addition, as part of our reordering analysis, we were able to discover Cross-Paragraph Sentence Reordering if they occurred in the same Sub-Document.
    \begin{enumerate}
        \item Paragraph Splitting
        \item Cross-Paragraph Sentence Reordering
        \item Paragraph Rearrangement
        \item Sub-Document Rearrangement 
        \item Adding Paragraphs
        \item Adding Sub-Documents
        \item Deleting Paragraphs
        \item Deleting Sub-Documents
    \end{enumerate}
\end{enumerate}


\section{Experiment and Results on the Manually-aligned Dataset}
\label{app:manual}

In this section, we describe the experimental setting and results for training TS models on a manually aligned dataset. We do so for completeness, since manually aligned datasets can potentially capture more complex relationships between source and target sentences than automatic alignments can, and the test dataset in CS is manually aligned. We report results for this series of experiments in an appendix, since no prior work used these datasets to train TS models.

The Manual dataset is created by combining WikiManual and Newsela Manual from \citet{jiang-etal-2020-neural}. 
\citet{jiang-etal-2020-neural} used WikiManual and NewselaManual to train their NeuralCRF sentence alignment models for the WikiLarge and Newsela corpora, respectively.
In addition, we use these datasets as other comparison points between TS and CS data presented in \S\ref{sec:sim_compare}.

With respect to SI counts, for the Manual dataset we use 1522/280 SIs from WikiManual and 11728/1418 SIs from NewselaManual to create combined training and validation sets of 11728/1418 SIs respectively. Although both WikiManual and NewselaManual contain tests sets that \citet{jiang-etal-2020-neural} used to test their CRF models, we use other datasets as the tests sets for our experiments (see \S\ref{sec:eval_data}).

We should note, that there are more SIs in the original datasets than the number of SIs we used for fine-tuning. This difference is because the missing SIs are either complete deletions (sentences from the source that are removed in the simplification) or complete additions (sentences in the simplification with no source). See \autoref{tab:si_counts} in \autoref{app:ops_analysis} for additional details regarding SI counts in each corpus.

\paragraph{Results.} When trained in this setting, which uses a considerably smaller albeit cleaner dataset, we notice two phenomena when compared to the results in \S\ref{sec:results} when tested on TS. First, for all models except T5Large, the +Classifier variant still outperforms the baseline model, though by a smaller margin than in the classic training setting. Second, model size now has a consistent trend, with larger models outperforming their matching smaller counterparts. Further work is required to ascertain this different pattern of performance on this setting.
In general, the best TS performance on SARI is achieved by the BART-Large+Classifier variant in this training setting, repeating the performance in \S\ref{sec:results}.

Examining the performance on CS, we find that the +Classifier variants achieved superior results for all model architectures and sizes in this training setting as well.
Unlike the results presented in \S\ref{sec:results}, here the difference in SARI scores is more pronounced for larger models, with differences of more than 1.3 SARI points for both large model architectures, while the differences in the base-sized models is under 0.8 SARI points. The model with the highest performance difference in this training setting is BART-Large+Classifier, with a difference of 5.05 SARI points on CS data, while in \S\ref{sec:results} this was the T5Base+Classifier model.

In both evaluation settings, the best performing model is still BART-Large+Classifier, similar to the results in \S\ref{sec:results}.

\paragraph{Discussion.} The results shown here further demonstrates the potential benefit of adding inductive bias towards simplification operations to a TS trained model. Potential future research could also look into performances of different models when trained on datasets of different sizes and quality, since many language lack resources for automatic text simplification, let alone cognitive simplification.

\section{SARI Calculation}
\label{app:SARI}
The main metric used for evaluating TS models is SARI \citep{xu-etal-2016-optimizing}, which is computed based on three token-level operations: ADD, KEEP, and DELETE. 
Precision and Recall are computed for each with respect to $n$-grams for $n=1\ldots 4$, and averaged together to yield overall Precision and Recall scores per operation. SARI is defined as:

\begin{equation}
    SARI=\frac{\textrm{F1}_{ADD}+\textrm{F1}_{KEEP}+\textrm{P}_{DELETE}}{3}
\end{equation}

\section{Model Training times}
\label{app:train_time}
\begin{table}[ht]
    \centering
    \begin{tabular}{c|c|c}
         \textbf{Train Dataset} & \textbf{Model Size} & \textbf{Train Time} \B\\ \hline
         \multirow{4}{*}{WikiAuto}& T5-Large & 7 days \T\\
         & T5-Base & 4 days \B\\
         \cline{2-3}
         & BART-Large & 5 days \T\\
         & BART-Base & 2 days \B\\ \hline
         \multirow{4}{*}{Manual}& T5-Large & 1 day \T\\
         & T5-Base & 12 hours \B\\
         \cline{2-3}
         & BART-Large & 20 hours \T\\
         & BART-Base & 11 hours \\
    \end{tabular}
    \caption{Approximate training times on a single GPU for our models trained in \S\ref{sec:experiments} and \autoref{app:manual}.}
    \label{tab:my_label}
\end{table}

\section{Comparing automatic identification of simplification operation to human annotations}
\label{app:token_ident_scores}
We asked a human annotator to manually assign simplification operations to 50 random SI from the WikiAuto training set. Below are the particular Precision, Recall, and F1 scores for each operation on that subset, using the human annotations as ground-truth.

\begin{table}[ht]
    \centering
    \small
    \begin{tabular}{c|c|c|c|c}
    Operation & P. & R. & F1 & \# \\
    \hline
    $<$\textsc{prox}$>$ & 0 & 0 & 0 & 0 \\
    $<$\textsc{rephrase}$>$ & 80.43 & 97.37 & 88.1 & 38 \\
    $<$\textsc{del}$>$ & 80 & 84.21 & 82.05 & 19 \\
    $<$\textsc{add}$>$ & 12.5 & 50 & 20 & 2 \\
    $<$\textsc{example}$>$ & 0 & 0 & 0 & 0 \\
    $<$\textsc{explain}$>$ & 0 & 0 & 0 & 0 \\
    $<$\textsc{explicit}$>$ & 42.86 & 42.86 & 42.86 & 7 \\
    $<$\textsc{reorder}$>$ & 32.43 & 1 & 48.98 & 12  \\
    $<$\textsc{split}$>$ & 1 & 1 & 1 & 13 \\
    \end{tabular}
    \caption{Precision, Recall, and F1 scores for each operation token, when comparing our automatic identification rules to a human annotator. We also describe the number of SI with each operation in the random sample analyzed, and the expected number SI.}
    \label{tab:sim_ops_human}
\end{table}

\section{Simplification Instance Counts}
\autoref{tab:si_counts} contains the details regarding the counts of SIs in each dataset, as used to fine-tune our models in \S\ref{sec:experiments}, and the full dataset, including deletions of complete sentences from the source and additions complete sentences to the target.

\begin{table}[ht]
    \small
    \begin{center}
    \begin{tabular}{c|c|c}
    \textbf{Dataset}     & \textbf{Fine-Tuning} & \textbf{Full Corpus} \B\\\hline
    \textbf{FA} & - / - / 321 & - / - / 380 \T\B\\ \hline
    \textbf{NewM} & 11.7K / 1.4K / 3.6K & 17.8K / 2.6K / 5.1K \T\B\\
    \textbf{WikiM} & 1.5K / 280 / 531 & 29.9K / 4.4K / 7.9K \B\\\hline
    \textbf{ASSET} & - / 2K / 359 & - / 2K / 359 \T\\
    \textbf{WikiA} & 483K / 20K / - & 483K / 20K / - \\
    \end{tabular}
    \end{center}
    \caption{Number of SIs used for fine-tuning our models in \S\ref{sec:results} and \autoref{app:manual} as compared to the number of SIs in the respective full corpus. The differences are because in the fine-tuning setting we ignored complete deletions of sentences from the source and complete additions of sentences to the target. For each dataset and each setting, the number of SIs are for the train / valid / test sets respectively. We shorten dataset names as follows: FA=FestAbility, NewM=NewselaManual, WikiM/A=WikiManual/Auto.}
    \label{tab:si_counts}
\end{table}

\begin{figure*}[tbhp]
\begin{subfigure}{\columnwidth}
    \centering\includegraphics[width=\columnwidth]{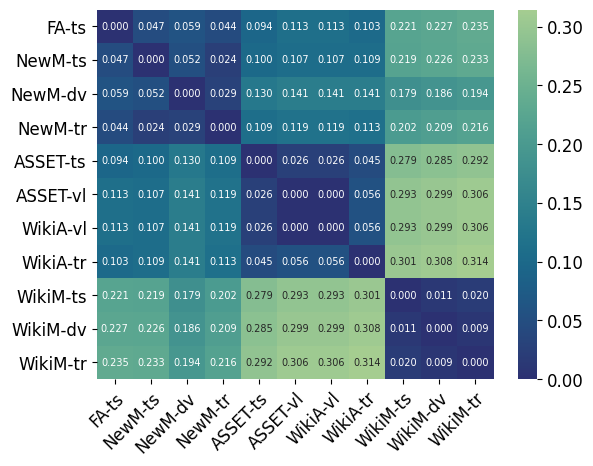}
    \caption{$\overline{JSD}$ distances between distributions
    }
    \label{fig:JSD_full}
\end{subfigure}
\hspace{10pt}
\begin{subfigure}{\columnwidth}
    \centering\includegraphics[width=\columnwidth]{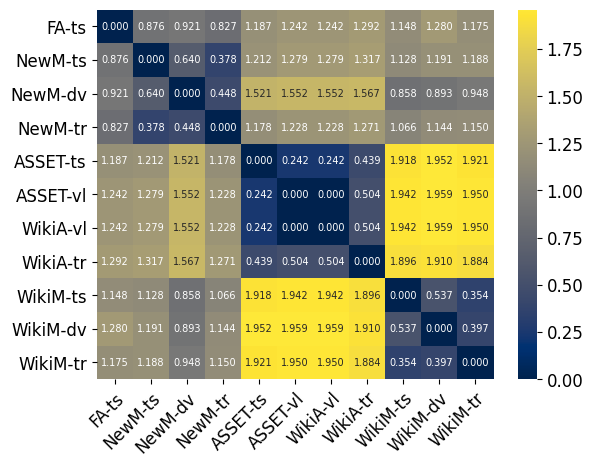}
    \caption{$\ell_2$ distances between  correlation matrices
    }
    \label{fig:l_2_full}
\end{subfigure} 
    \caption{
    Heatmaps of the distances between dataset sub-sets. We abbreviate sub-set names such that FA=FestAbility, NewM=NewselaManual, WikiM/A=WikiManual/Auto. The final two letters signify ts=test, vl=valid, dv=dev, and tr=train sets. For each sub-set pair, we report the numerical distance in the matching cell.
    }
    \label{fig:full_corpora}
\end{figure*}

\section{
Simplification Operations per Dataset}
\label{app:sim_ops_details}
\begin{figure*}[tbhp]
    \centering
    \includegraphics[width=\textwidth]{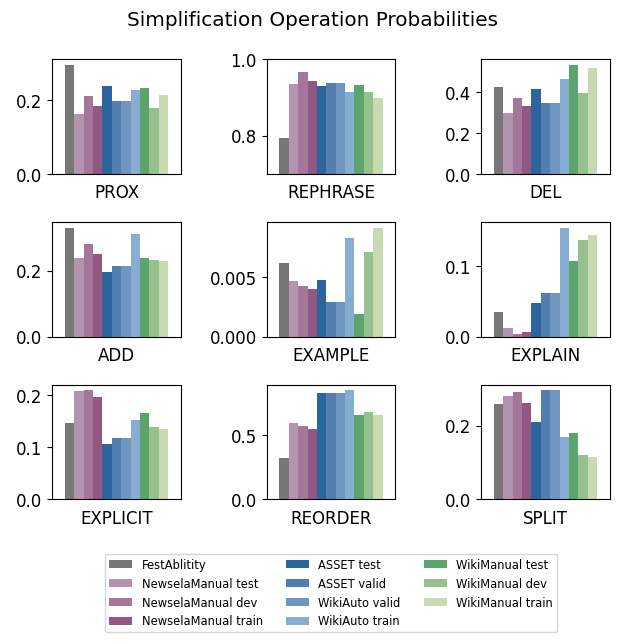}
    \caption{Probabilities each simplification operation is used in every dataset sub-set}
    \label{fig:probabilitis}
\end{figure*}

\begin{figure*}[tbhp]
    \centering
    \includegraphics[width=\textwidth]{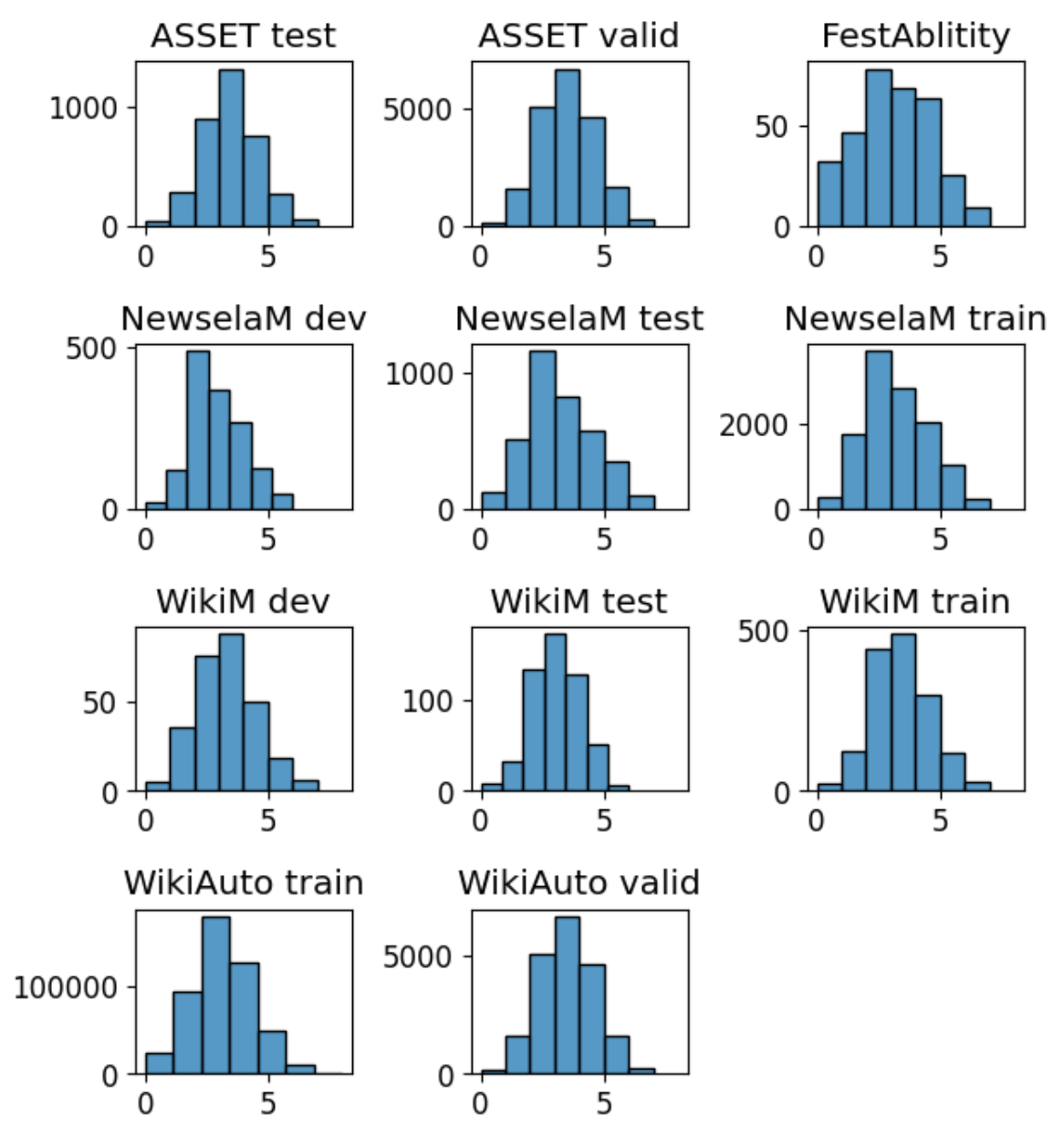}
    \caption{Histograms of the number of simplification operations used in each SI for each dataset sub-set.}
    \label{fig:histograms}
\end{figure*}

\begin{figure*}[tbhp]
    \centering
    \includegraphics[width=\textwidth]{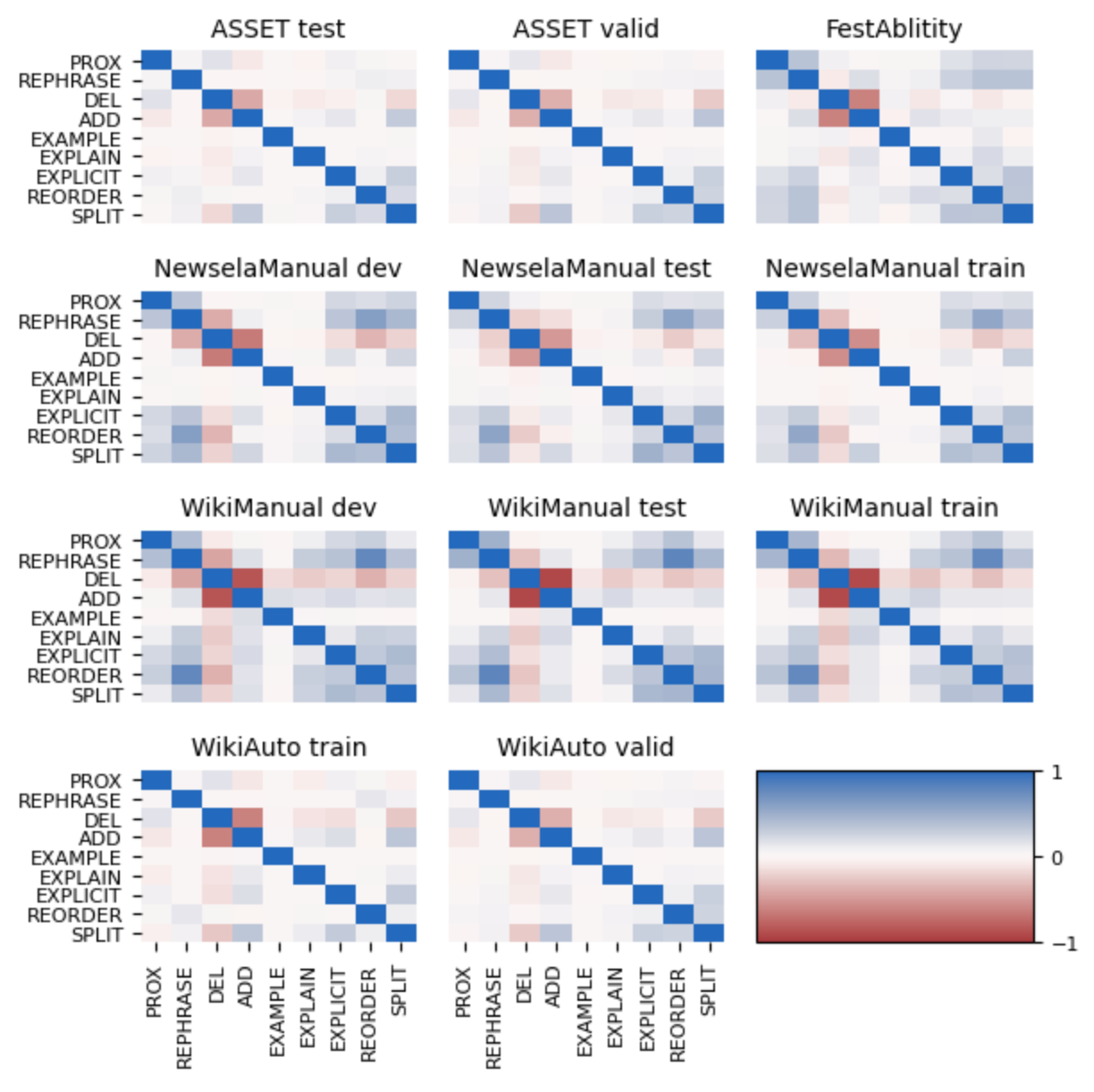}
    \caption{The simplification operations correlations matrices for each dataset subset presented as heatmaps.}
    \label{fig:correlations}
\end{figure*}

In this appendix, we present the results of 3 key point of information regarding the use of simplification operations in the TS and CS datasets.
First, we show the distribution of each simplification operations per dataset (\autoref{fig:probabilitis}). Then, we show the histograms of the number of simplification operations used in each SI (\autoref{fig:histograms}). Finally, we present the correlation matrices for each dataset used in our analysis in \S\ref{sec:sim_compare} (\autoref{fig:correlations}).


\section{Full Corpora Analysis}\label{app:ops_analysis}
In the main paper \S\ref{sec:sim_compare}, we analyzed simplification operations in the datasets as they were used to train our models. However, each dataset also has SIs that are complete deletions (whole sentences in the source that don't have matching sentence(s) in the target) or complete additions (sentences in the target with no matching source sentence(s)). In figure \autoref{fig:full_corpora} we present the results of the same analysis but for the full dataset.

When analyzing the full datasets, similar patterns to \S\ref{sec:sim_compare} emerge. Sub-sets of the same dataset are still clustered together, although now the in-cluster distance is $\overline{JSD}<0.06$ and between clusters distance is $\overline{JSD}>0.09$. Moreover, if considering clusters to have $\overline{JSD}<0.04$ like in the main paper, then the similar relationships between sub-sets described in the main paper emerge -- FestAbility is clustered with itself, ASSET and WikiAuto validation are clustered, and WikiManual is also clustered, and WikiAuto train is separated from the ASSET and WikiAuto validation cluster and the WikiManual cluster. However, there are some differences -- NewselaManual development is not clustered with the other NewselaManual sub-sets if considering clusters to have $\overline{JSD}<0.04$, and WikiAuto train is not closer to the WikiManual cluster than to the ASSET and WikiAuto validation cluster.

These results strengthen our findings from the main paper that the simplification operations are used similarly in CS and TS. They also emphasize the differences between the Newsela corpus and the WikiLarge corpus, as highlighted by \citet{xu-etal-2015-problems}. The difference between WikiManual and all the other datasets is the prevalence for ``full deletions'' in WikiManual, which shows that the relationship between English Wikipedia and Simple English Wikipedia contains many more cases of Information Deletion than other corpora.

In addition, the distances between the operation correlation matrices show that the difference in joint application of simplification operations between CS and TS is similar when considering the full datasets, as the distances between FestAbility and the other sub-sets are maintained (changing by at most $\pm0.26$, while the other distances outside of clusters increase more).

\section{Example Simplifications}
Shown in \autoref{tab:examples_asset} and \autoref{tab:examples_CS} below.

\begin{table*}[hb]
    \centering
     \renewcommand{\arraystretch}{1.5}
    \begin{tabular}{lp{0.7\linewidth}}
        \hline
         \textbf{Source}: & Jeddah is the \textbf{principal} gateway to Mecca, Islam's holiest city, \textit{which able-bodied Muslims are required to visit at least once in their lifetime.} \\
         \textbf{Simplification}: & Jeddah is the \textbf{main} gateway to Mecca, Islam's holiest city. \\
         \hline
         \textbf{Source}: & However, the BZ differs a bit \textit{in comparison} to the Freedom Party, as is in favor of a referendum about the Lisbon Treaty but against an EU-Withdrawal. \\
         \textbf{Simplification}: & However, the BZ differs a bit from the Freedom Party\textbf{. The BZ} is in favor of a referendum about the Lisbon Treaty but against an EU-Withdrawal. \\
         \hline
         \textbf{Source}: & Many species had \textbf{vanished} by the end of the nineteenth century\textit{, with European settlement}. \\
         \textbf{Simplification}: & Many species had \textbf{disappeared} by the end of the nineteenth century. \\
         \hline
         \textbf{Source}: & \textbf{Fearing that Drek will destroy the galaxy, Clank asks Ratchet} to help him find the famous superhero Captain Qwark,\textit{ in an effort to stop Drek}. \\
         \textbf{Simplification}: & \textbf{Clank fears that Drek will destroy the galaxy. He asks Ratchet} to help him find the famous superhero Captain Qwark.\\
         \hline
         
    \end{tabular}
    \caption{Example Simplifications from ASSET of the T5-Base Classifier model fine-tuned on the WikiAuto dataset. Differences between the source and Simplification are bolded and italicized per example.}
    \label{tab:examples_asset}
\end{table*}

\begin{table*}[hb]
    \centering
     \renewcommand{\arraystretch}{1.5}
    \begin{tabular}{lp{0.7\linewidth}}
        \hline
         \textbf{Source}: & Know that there are absentee ballot options available, \textit{and there may be other options available depending on what situation we find ourselves in}. \\
         \textbf{Simplification}: & There are \textbf{many} options for absentee ballots. \\
         \textbf{Reference:} & You should know if you could vote from home. You should check if there are other ways of voting this year because of covid-19.\\
         \hline
         \textbf{Source}: & Zazel O'Garra\textbf{, founder} and artistic director of ZCO Dance Project, \textit{is a force to be reckoned with}. \\
         \textbf{Simplification}: & Zazel O'Garra \textbf{is the founder} and artistic director of ZCO Dance Project. \\
         \textbf{Reference:} & Zazel O'Garra is the founder and artistic director of ZCO Dance Project. She is a very strong and important woman.\\
         \hline
         \textbf{Source}: & \textit{I was diagnosed at the age of five} and completely stopped walking when I was 12 years old. \\
         \textbf{Simplification}: & I stopped walking when I was 12 years old. \\
         \textbf{Reference:} & It started when I was five years old. I stopped walking when I was 12 years old. \\
         \hline
         \textbf{Source}: & \textit{And in trying to get towards where I really wanted to be in life}, I really thought about giving back and being of service to someone else. \\
         \textbf{Simplification}: & I really thought about giving back and being of service to someone else.\\
         \textbf{Reference:} & I thought about what I really wanted to do. I wanted to help other people. \\
         \hline
         
    \end{tabular}
    \caption{Example simplifications from FestAbility Transcripts of the T5-Base Classifier model fine-tuned on the WikiAuto dataset. Differences between the source and simplification are bold-faced and italicized, per example. We add the CS reference for each example as well to highlight where the model succeeds and fails in the task.}
    \label{tab:examples_CS}
\end{table*}

\end{document}